\documentclass[a4paper]{article}
\usepackage{INTERSPEECH2019}

\title{SANTLR: Speech Annotation Toolkit for Low Resource Languages}
\name{Xinjian Li, Zhong Zhou, Siddharth Dalmia, Alan W. Black, Florian Metze}
\address{Language Technologies Institute, Carnegie Mellon University; Pittsburgh, PA; U.S.A.
\email{\{xinjianl|zhongzho|sdalmia|awb|fmetze\}@cs.cmu.edu}}

\begin{document}

\maketitle
\begin{abstract}
While low resource speech recognition has attracted a lot of attention from the speech community, there are a few tools available to facilitate low resource speech collection. In this work, we present SANTLR: Speech Annotation Toolkit for Low Resource Languages. It is a web-based toolkit which allows researchers to easily collect and annotate a corpus of speech in a low resource language. Annotators may use this toolkit for two purposes: transcription or recording. In transcription, annotators would transcribe audio files provided by the researchers; in recording, annotators would record their voice by reading provided texts. We highlight two properties of this toolkit. First, SANTLR has a very user-friendly User Interface (UI). Both researchers and annotators may use this simple web interface to interact. There is no requirement for the annotators to have any expertise in audio or text processing. The toolkit would handle all preprocessing and postprocessing steps. Second, we employ a multi-step ranking mechanism facilitate the annotation process. In particular, the toolkit would give higher priority to utterances which are easier to annotate and are more beneficial to achieving the goal of the annotation, e.g. quickly training an acoustic model. 




\end{abstract}

\section{Introduction}

In recent years, speech recognition has witnessed a lot of progress by successfully applying deep learning \cite{xiong2016achieving}. While rich resource languages such as English and Mandarin have benefited enormously from this progress, research in low resource languages suffer from the lack of speech training data. Ethonologue, which is one of the most extensive catalog of the world's language, has estimated the total number of languages in this world is about 7000~\cite{lewis2009ethnologue}. However, only a small portion of those languages have audio or language resources, not to mention clean parallel speech training corpus. For example, Bible, the most translated text, has only been translated into 2508 languages as of 2009~\cite{lewis2009ethnologue}.

While there is an increasing interest in building speech recognition systems for low resource languages, few annotation toolkits are available in the community to collect speech data from native speakers, or they are difficult to use\cite{schultz2007spice}. In this work, we present our SANTLR system: Speech Annotation Toolkit for Low Resource Languages. The toolkit mainly implements two applications: \textbf{transcribe} application and \textbf{record} application. The first application is for the annotators to transcribe texts from audios, the other one is for the annotators to record their voice by providing the reading text. We designed our toolkit to have a very simple user interface for both researchers and annotators, which we would briefly introduce in the next section. Then in the section 3, we describe our utterance ranking system in SANLTR which particularly aims to take full advantange of annotator's time. In section 4, we show some actual statistics we collected in the previous LOREHLT evaluations~\cite{chaudhary2019ariel}. We name the toolkit after the famous LL parser ANTLR, and hope it could be widely used in the speech community.

\section{User Interface}
The first highlight of this toolkit is its user interface. We designed our user interface so that it is easy to use for both researchers and annotators. To start with, researchers would first be asked to upload either audios or texts. Those audios and texts might be noisy or unstructured. Then the toolkit could automatically handle all the preprocessing steps of audios and texts. For example, it would clean up all text files by removing HTML tags or emojis, it would also split long audios into small ones by voice activity detection. This feature enables researchers to easily collect speech data even without any expertise in the realm of audio processing or natural language processing. 

After waiting for the preprocessing step to finish, a link would be generated for this specific annotation task. The link is accessible from both the researcher side and annotator side. Researchers can simply share the link to annotators, then ask them to start their annotation task. The link can be shared by multiple annotators to perform the annotation task simultaneously. Furthermore, researchers could monitor their progress with the same link. Regarding the annotator side, our applications are built with recent web front-end framework and they are accessible from various devices such as desktop computers and mobile phones. It also supports multiple user-friendly features. For example, in the transcribe application, each audio would be streamed as a mp3 audio and would be available to the annotator in a lazy loading manner. This helps to increase the loading speed and reduce network traffics. Additionally, we implemented the auto-saving feature so that annotators do not need to worry about forgetting saving their annotations during the tasks. After the task finished, researchers can download both audios and its corresponding transcriptions by using our web interface. 

\section{Utterance Ranking}
Our novelty in this toolkit is our multi-step utterance ranking mechanism. Most existing annotation tools require native speakers to annotate audios/texts in a sequential manner, this strategy works well for rich resource languages as the annotators' cost is relatively low and it could obtain various speech training data. We argue, however, this sequential annotation approach is not a good option in the case of low resource languages because the annotation time is highly limited. In such a case, we should give different priority to each utterance by considering their characteristics so that we could take full advantage of annotator's valuable time. For example, some utterances might be easier to annotate and more beneficial to the acoustic model, therefore those utterances should be given higher priority to annotate. On the other hand, noisy audios which contain less speech data should be given lower priority because they are hard to annotate and to be learnt from. For this reason, we implement the utterance ranking system within the toolkit to tackle this problem. We describe our strategy of both the audio ranking and text ranking in the following part.

\subsection{Audio Ranking}

For transcription, our ranking mechanism has three steps. The first step is to sort audios based on their durations in the ascending order. Our previous annotation experiment has suggested that transcribing short audios is a much easier task for annotators when compared with longer audios. The reason is that longer audios would implicitly require annotators to listen to the entire audios repeatedly so that they could remember the entire contents and then start to transcribe. Additionally, they are prone to make more mistakes when transcribing longer audios. On the other hand, shorter audios are easier to transcribe and one-time listening might be enough in some cases. In addition to the annotator's cost, our preference of shorter audios could also be justified from the training perspective, since it is easier for acoustic model to learn alignments between phonemes and frames when audios are shorter. There is also research work which empirically proves shorter training data would benefit the acoustic model~\cite{amodei2016deep}. Therefore, we first sort all audios by their length. Next, we compute the S/N ratio of each audio and adjust our previous ranking based on its S/N ratio. The intuition of this step is to remove noisy and unclear audios from the training set.

The third step is to rank audios based on their phoneme overlaps. This step is intended to increase the variety of training audios. During our experiment, we noticed that there were many duplicate utterances in both audios and texts side. Those duplicate utterances should be removed as they provide few new information to the acoustic model and it might even cause the model to overfit. For instance, we might end up with a lot of training utterances of \textit{yeah} or \textit{no} in English annotations. Obviously those are not very useful to train a good English model. As the target low resource language usually has no training data, we estimate their phonemes by using a pretrained EESEN English acoustic model~\cite{miao2015eesen}. For each audio after the two previous ranking steps, we compute its phoneme overlap with every higher ranked audio. If we detect that it has a high overlap, then we decrease its current ranking based on its overlap score as we do not want annotators to transcribe similar audios twice.

\subsection{Text Ranking}

For recording, we have a similar ranking system as the one used in transcription. There are two steps to rank texts. First, a small language model is estimated by using all the provided texts. Next we use the language model to compute perplexity of each sentence. The perplexity is normalized by the length of each sentence. The first sorting would be done based on the perplexity score in the ascending order. This aims to ask annotators to read texts which contains frequent words rather than rare words. Second, we filter the duplicate texts in a similar way as we perform in the audio rankings. We compute the overlap between two texts by computing the edit distance between them. A high overlap would decrease its ranking to allow annotators to read more diverse texts. We do not, however, penalize texts based on its length or the number of words. This is because it is a much easier task to read rather than transcribe. We did not observe any negative effects of reading longer texts in our experiment.

\section{Experiments}

The SANTLR toolkit has been developed for ARIEL-CMU systems in LOREHLT 2018~\cite{chaudhary2019ariel}. It was actually deployed to collect speech training data from multiple native speakers for several languages, as shown in Table~\ref{tab:stats}. For each language, we show the average number of words and the audio duration we collected per hour. Each annotator was typically asked to transcribe for the first 30 minutes and then record their voice in the remaining 30 minutes. 

\begin{table}[h!]
\begin{center}
\caption{\small Statistics of speech data collected per hour.}
\begin{tabular}{l r r}
\hline
    \textbf{Language} & \textbf{number of words \#} & \textbf{audio minutes \#} \\ \hline
    Thai & 648 & 9.6 \\
    Hindi & 277 & 4.2 \\
    Kinyarwanda & 1,044 & 8.2 \\
    Sinhala & 820 & 6.2 \\ \hline
\end{tabular}
\label{tab:stats}

\end{center}
\end{table}

We found that in general annotators who had high computer literacy would perform those annotation tasks much faster and they could understand the annotation interface very easily. Otherwise, we need to prepare our annotators with more user instructions. 

\section{Conclusions}
In this paper, we present our annotation toolkit SANTLR. We describe the most important aspects of the user interface, and the ranking strategy, to allow for efficient annotation. The toolkit would be released to Github under an open source license soon to benefit the research community.

\section{Acknowledgements}
This project was sponsored by the Defense Advanced Research Projects Agency (DARPA) Information Innovation Office (I2O), program: Low Resource Languages for Emergent Incidents (LORELEI), issued by DARPA/I2O under Contract No. HR0011-15-C-0114. 

\bibliographystyle{IEEEtran}

\bibliography{mybib}


\end{document}